\documentclass[conference]{IEEEtran}
\IEEEoverridecommandlockouts

\usepackage{cite}
\usepackage[hidelinks,bookmarks=false]{hyperref}
\usepackage{amsmath,amssymb,amsfonts}
\usepackage{algorithmic}
\usepackage{graphicx}
\usepackage{textcomp}
\usepackage{xcolor}
\def\BibTeX{{\rm B\kern-.05em{\sc i\kern-.025em b}\kern-.08em
    T\kern-.1667em\lower.7ex\hbox{E}\kern-.125emX}}

\usepackage{booktabs} 
\usepackage{array}
\usepackage{url}
\usepackage[font=footnotesize]{caption}
\usepackage{setspace}
\usepackage{latexsym}
\usepackage{svg}
\usepackage{float}
\usepackage{stfloats}

\usepackage[switch]{lineno}
\usepackage{pifont} \usepackage{xspace}

\usepackage{bbding}
\usepackage{wrapfig}

\newif\iffinal

\definecolor{darkpastelgreen}{rgb}{0.01, 0.75, 0.24}
\def\yes{\textcolor{darkpastelgreen}{\Checkmark}\xspace}
\def\no{\textcolor{red!20}{\XSolidBrush}\xspace}

\usepackage{xcolor}
\usepackage{tikz}
\usepackage{amsmath}
\usepackage{amssymb}
\usepackage{hyperref}

\definecolor{anchorcolor}{RGB}{3,101,192}
\definecolor{positivecolor}{RGB}{81,167,249}
\definecolor{negativecolor}{RGB}{236,93,87}

\newcommand{\circleanchor}[1]{%
  \begin{tikzpicture}[baseline=(char.base)]
    \fill[fill=anchorcolor] (0,0.025) circle (0.2cm);
    \node[text=white, font=\bfseries\scriptsize] (char) at (0,0.025) {#1};
  \end{tikzpicture}%
}

\newcommand{\circlepositive}[1]{%
  \begin{tikzpicture}[baseline=(char.base)]
    \fill[fill=positivecolor] (0,0.025) circle (0.2cm);
    \node[text=white, font=\bfseries\scriptsize] (char) at (0,0.025) {$#1$};
  \end{tikzpicture}%
}

\newcommand{\circlenegative}[1]{%
  \begin{tikzpicture}[baseline=(char.base)]
    \fill[fill=negativecolor] (0,0.025) circle (0.2cm);
    \node[text=white, font=\bfseries\scriptsize] (char) at (0,0.025) {$#1$};
  \end{tikzpicture}%
}
\finaltrue

\IEEEoverridecommandlockouts
\IEEEpubid{\makebox[\columnwidth]{979-8-3195-0255-1/26/\$31.00~\copyright2026 IEEE \hfill}
\hspace{\columnsep}\makebox[\columnwidth]{ }}

\begin{document}

\title{An overview of 3D Vision-Language Models}

\newcommand{\cmtid}{266}

\iffinal
\author{
\IEEEauthorblockN{
Márcus Lobo\textsuperscript{*}\IEEEauthorrefmark{4}\thanks{This work was supported by the São Paulo Research Foundation
    (FAPESP), grants \#2024/09462-1 and \#2026/01721-3, and by the
    Conselho Nacional de Desenvolvimento Científico e Tecnológico
    (CNPq), grant \#315158/2023-9.}, 
Vitor Matias\textsuperscript{*}\IEEEauthorrefmark{4}, 
Afonso Paiva\IEEEauthorrefmark{4},
Jeová Farias\IEEEauthorrefmark{3},
Tiago Novello\IEEEauthorrefmark{2}, and
Moacir Ponti\IEEEauthorrefmark{4}}
\IEEEauthorblockA{
\textit{ICMC-USP\IEEEauthorrefmark{4}, IMPA\IEEEauthorrefmark{2}, Bowdoin College\IEEEauthorrefmark{3}}\\
}
}

\maketitle
\begingroup
\renewcommand\thefootnote{*}
\footnotetext{denotes equal contribution.}
\endgroup

\else
  \author{
\IEEEauthorblockN{SIBGRAPI Paper ID: \cmtid} }
  \linenumbers

  \maketitle
\fi


\maketitle

\begin{abstract}
Vision-Language Models (VLMs) are reshaping computer vision by aligning visual and textual embeddings, allowing models to recognize visual concepts and reason about them using natural language.
Traditional 3D deep-learning models, however, are typically trained for specific tasks, such as classification, segmentation, or detection, and do not naturally support cross-modal retrieval from their embedding spaces using text or images as queries.
To address this issue, Contrastive Language-Image Pretraining (CLIP)-based methods align 3D embeddings with pretrained image and text representations, giving rise to 3D Vision-Language Models (3D VLMs) that support zero-shot classification and cross-modal retrieval, and open-vocabulary recognition of 3D shapes.
This tutorial provides an overview of 3D VLMs, ranging from basic definitions of 3D representations and their encoding into embeddings to cross-modal contrastive alignment, modern multimodal frameworks, and 3D Vision-Large Language Models (3D VLLMs). We present the main definitions of contrastive learning for multimodal embedding alignment and highlight recent advances in language-guided 3D Gaussian splatting, 3D shape generation, and embodied AI for robotics. Tutorial page: \url{https://usmarcv.github.io/Tutorial-3DVLMs/}.
\end{abstract}


\section{Introduction}
\label{sec:introduction}

Understanding 3D objects and scenes requires jointly reasoning about geometry, appearance, and semantics. Unlike images, which are regular pixel grids, 3D data can be represented as point clouds, meshes, multi-view images, sparse voxels, and 3D Gaussian primitives. Each representation exposes different geometric and visual cues and requires specialized encoding strategies.
Traditional 3D learning has primarily focused on closed-set recognition, where a model maps an input shape or scene to a category from a fixed taxonomy shared by training and inference. Methods such as MVCNN~\cite{su2015multi} and PointNet~\cite{Qi2017pointnet} established architectures for multi-view images and point clouds, respectively. However, their output spaces are typically defined by task-specific class labels and are not directly connected to natural-language semantics. 

\begin{figure}[h!]
    \centering
    \includegraphics[width=1\linewidth]{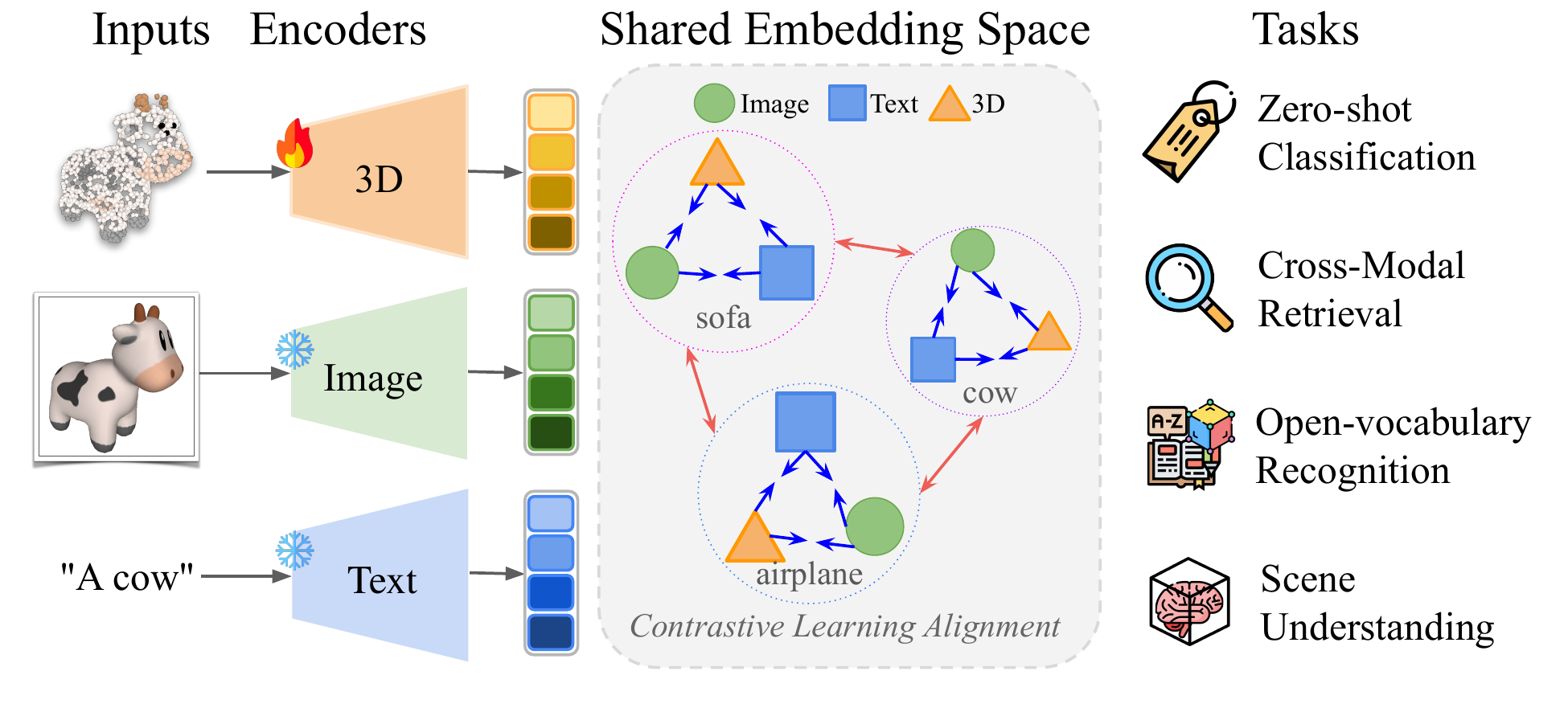}
    \caption{\textbf{Overview of the 3D VLM pipeline.} Cross-modal contrastive objectives increase the similarity of matched 3D--Image and 3D--text pairs relative to negative in-batch candidates, producing a shared semantic embedding space.
    }
    \label{fig:overview}
\end{figure}

Vision-Language pre-training provides a path beyond this limitation.
CLIP~\cite{Radford2021learning} learns a shared embedding space in which corresponding images and texts have high similarity. \textbf{3D} \textbf{V}ision-\textbf{L}anguage \textbf{M}odels (3D VLMs) extend this principle by aligning geometric representations with pretrained image and text embeddings, as shown in \autoref{fig:overview}. Given a 3D object or scene, its 2D visual observations, and associated language, these models learn representations that support zero-shot classification, open-vocabulary comprehension, and cross-modal retrieval\footnote{\textbf{Zero-shot classification}: categorizes samples into unseen classes without label-specific training. \textbf{Open-vocabulary comprehension}:  ability to recognize and reason about arbitrary semantic concepts defined via natural language. \textbf{Cross-modal retrieval} enables searching across modalities, such as retrieving a 3D shape given a query containing text or a 2D image.}~\cite{Xue2023ulip, hegde2023clip, Liu2023openshape}.
We distinguish these embedding-based models from \textbf{3D V}ision \textbf{L}arge \textbf{L}anguage \textbf{M}odels (3D VLLMs). While 3D VLMs primarily produce comparable representations or similarity scores, 3D VLLMs map 3D features or tokens into a large language model, enabling captioning, question answering, instruction following for embodied AI and agentic workflows, and spatial reasoning\footnote{\textbf{Question answering}: answering natural language queries about 3D scenes. \textbf{Instruction following}: executing complex tasks or actions based on user text commands. \textbf{Spatial reasoning}: understanding dynamic spatial relationships, relative layouts, and orientations within 3D environments.}.

\begingroup
\setlength{\columnsep}{2pt}%
\setlength{\intextsep}{2pt}

Vision-language modeling has attracted increasing attention, with related work accounting for roughly 25\% of CVPR 2026 papers, as shown in the figure below~\cite{Numberofpaperscvpr}.    
\begin{wrapfigure}{r}{0.32\columnwidth}
\centering
    \includegraphics[width=0.3\columnwidth]{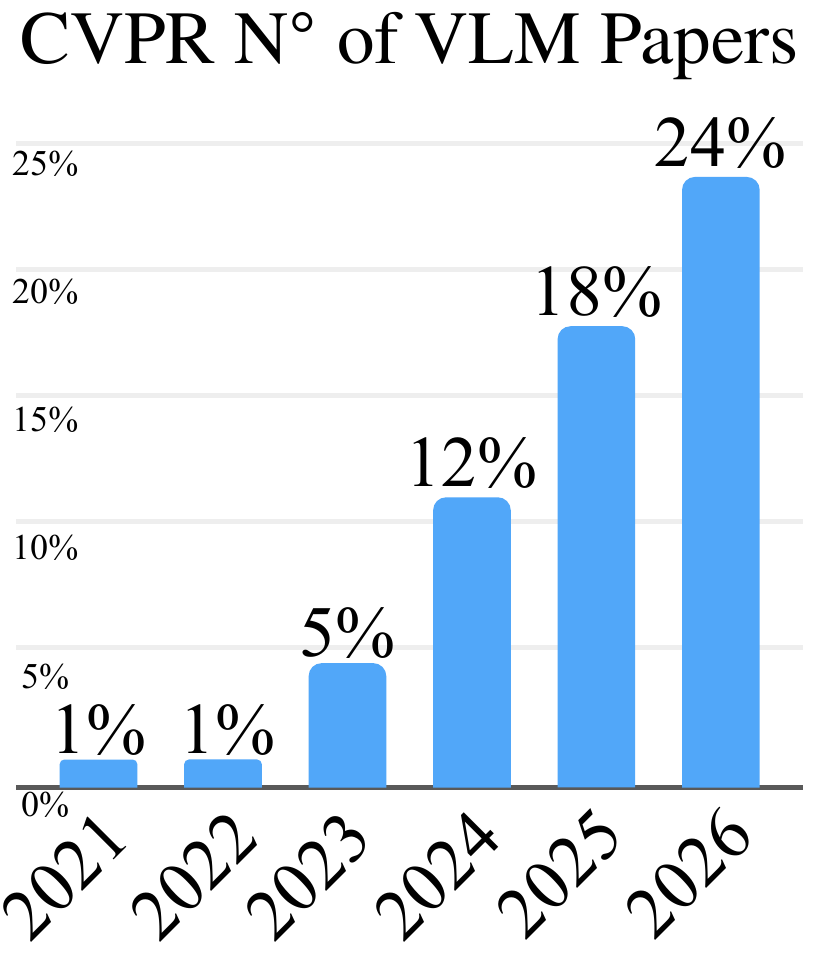}
\end{wrapfigure}
Existing surveys, as summarized in \mbox{\autoref{tab:related_surveys}}, tried to cover this area, but they typically focus on general vision-language pretraining, point cloud foundation models, 3D multimodal language models, language-grounded scene understanding, or multimodal generation. These studies provide extensive coverage of their respective topics but do not jointly connect 3D representations, CLIP-style alignment, embedding-based 3D VLMs, 3D VLLMs, and language-aligned 3D Gaussian representations.

\begin{table}[t]
\centering
\caption{Scope comparison with related surveys. \yes: dedicated coverage; $\circ$: partial or task-specific coverage; \no: outside the primary scope. \emph{Rep.}: multiple 3D representations; \emph{CL}: contrastive learning. }
\label{tab:related_surveys}
\setlength{\tabcolsep}{5pt}
\renewcommand{\arraystretch}{0.2}
\resizebox{\columnwidth}{!}{
\begin{tabular}{lccccc}
        \toprule
        Method &
        Rep. &
        CL &
        3D VLMs &
        3D VLLMs &
        3DGS/Applications \\
        \midrule
        Zhang et al.~\cite{zhang2024vlmsurvey}
        & \no & \yes & \no & \no & \no \\

        Ma et al.~\cite{ma2024llms}
        & \yes & $\circ$ & $\circ$ & \yes & $\circ$ \\

        Thengane et al.~\cite{thengane2025foundational}
        & $\circ$ & \yes & \yes & \yes & \no \\

        Ren et al.~\cite{ren2025languagegrounded}
        & $\circ$ & $\circ$ & $\circ$ & \yes & \no \\

        Li et al.~\cite{li2026embodied}
        & $\circ$ & $\circ$ & $\circ$& \yes & \no \\

        Guo et al.~\cite{guo2026visualgrounding}
        & $\circ$ & $\circ$ & $\circ$ & $\circ$ & $\circ$ \\

        Sapkota et al.~\cite{sapkota2025review}
        & $\circ$ & $\circ$ & \yes & $\circ$ & \no \\

        Hu et al.~\cite{hu2026simulation}
        & $\circ$ & $\circ$ & $\circ$ & $\circ$ & \yes\\

        \midrule
        \textbf{Ours}
        & \yes & \yes & \yes & \yes & \yes \\
\bottomrule
\end{tabular}
}
\end{table}

The goal of this tutorial is to present 3D VLM-based methods and trace their evolution from closed-set 3D recognition to open-world understanding. 
We then review the main 3D representations and the contrastive learning principles, definitions, and fundamentals behind cross-modal alignment. We then organize existing approaches into projection-based methods, joint embedding models, 3D Gaussian Splatting-based frameworks, and 3D Large Language Models. Finally, we discuss emerging applications and open challenges. In this tutorial, the participants should be able to:
\begin{itemize}
    \item Identify the main 3D representations and encoders used in multimodal learning;
    \item Explain CLIP-style alignment between 3D, image, and text embeddings;
    \item Distinguish embedding-based 3D VLMs from generative 3D VLLMs and select suitable models for different downstream tasks.
    
\end{itemize}
\endgroup

\section{Representations and Cross-Modal Alignment}
\label{sec:foundations}
\begin{figure}[htpb]
    \centering
    \includegraphics[width=1\linewidth]{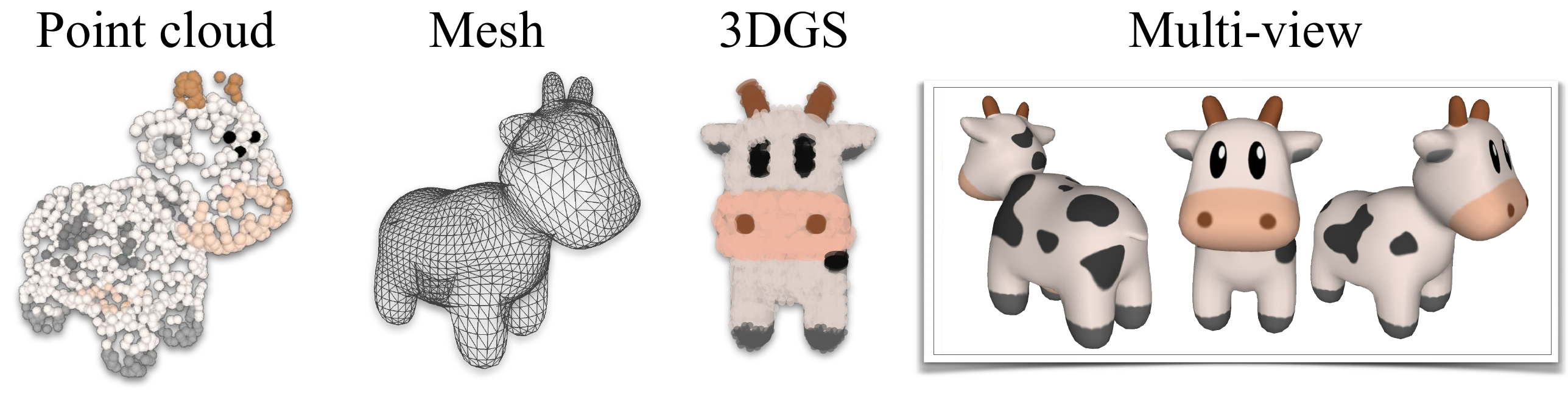}
    \caption{Examples of geometric representations used by 3D VLMs: point cloud, meshes, 3D Gaussian Splatting, and multi-view renderings.}
    \label{fig:3d-representations}
    \vspace{-0.5cm}
\end{figure}

\noindent \textbf{Problem setting.}~3D VLMs bridge inputs with fundamentally different structures, including geometry, images, and natural language. Given a dataset of $\mathcal{N}$ triplets $\{(S_i,I_i,T_i)\}_{i=1}^{{N}}$, where $\mathcal{S}=\{S_i\}$ denotes a 3D representation of an object, $\mathcal{I}=\{I_i\}$ is a 2D image (that could be obtained by rendering the 3D representation from an arbitrary viewpoint), and $\mathcal{T}=\{T_i\}$ is a textual description of the object.
Dedicated encoders, $E_S$, $E_I$, and $E_T$, process the 3D representation, images, and text descriptions, respectively. 
The architecture of $E_{S}$ depends on the geometric representation being processed, such as point clouds, multi-view images, meshes, or 3D Gaussian Splatting, see \autoref{fig:3d-representations}. In CLIP-based methods, $E_I$ and $E_T$ are commonly initialized from a pretrained vision-language model such as CLIP and may remain frozen, while $E_S$ is optimized to match its visual-semantic space.
In this case, text provides its semantic meaning, appearing as class names, object descriptions, captions, questions, referring expressions, or instructions (actions). 
These features are mapped by projection layers $P_{S}$, $P_I$, and $P_T$ into $\ell_2$-normalized embeddings $z^{S}$, $z^{I}$, and $z^{T}$ in a common embedding space $\mathbb{R}^{d}$.

\subsection{3D Inputs and Encoders}
\label{subsec:3d-representations}
Fundamentally, a 3D understanding application requires the choice of a surface representation.
Unlike 2D images, which follow a regular pixel grid, 3D data can be represented through multiple structures with different geometric, appearance, and computational properties. Each representation determines the encoder architecture used to extract the 3D features.
%
%
\noindent \textbf{Multi-view images:} 2D renders of a 3D object captured from different viewpoints $x_i^{I}={v_i^{(1)},\ldots,v_i^{(V)}}$, where $v_i^{(k)}$ denotes the $k$-th view of the object. Each rendering is processed by a shared 2D image encoder, benefiting from 2D foundational models backbones, typically a convolutional network, like MVCNN~\cite{su2015multi}, or a Vision Transformer, and the view-level features are aggregated into an object-level representation.
%
\noindent A \textbf{point cloud} is a set of 3D points $\{\mathbf{p}_i \in \mathbb{R}^3\}_{i=1}^N$, optionally with normals or colors~\cite{Qi2017pointnet}, and it is the native output of LiDAR scanners and depth sensors, making it the most direct representation for real-world 3D data. 
Their irregular and permutation-invariant structure motivates specialized architectures, including point-wise networks such as PointNet~\cite{Qi2017pointnet}, graph-based encoders such as DGCNN~\cite{wang2019dynamic}, and transformer-based models such as Point-BERT~\cite{Yu2022pointbert} and Point-MAE~\cite{pang2022masked}, or Point Transformer~\cite{Zhao2021pointtransformer} which constructs a self-attention network using vector self-attention and trainable 3D relative position encodings within local neighborhoods to adaptively aggregate spatial point features. On the other hand, SparseConv~\cite{Choy20194dsparseconv} is able to process high-dimensional, sparse spatio-temporal data directly by extending standard convolutions to generalized sparse tensor operations with custom kernel shapes. These encoders differ mainly in how they capture local neighborhoods and long-range geometric dependencies. Beyond point cloud encoding, recent methods decompose point clouds into compact, interpretable primitives~\cite{superflex}.

\noindent \textbf{Meshes:} A polygonal mesh defines an object's surface through triangles as vertices, edges, and faces, and is the standard representation in computer graphics. However, their connectivity poses challenges for learning operations. Early mesh-specific architectures include MeshCNN~\cite{hanocka2019meshcnn}, which defines convolution and task-driven pooling operations directly over mesh edges while preserving the underlying surface topology. MeshNet~\cite{feng2019meshnet} provides a complementary face-based formulation, representing each triangular face through spatial and structural features for shape classification and retrieval. 

\noindent \textbf{Implicit fields}: Neural implicit representations model surfaces as continuous occupancy~\cite{mescheder2019occupancy} or signed-distance fields~\cite{park2019deepsdf, novello:i3d:2022, novello2023neural, schirmer2024geometric}. Michelangelo aligns such representations with image and language embeddings using CLIP~\cite{zhao2023michelangelo}.

\noindent \textbf{3D Gaussian Splatting (3DGS):} Represents a scene as a collection of anisotropic 3D Gaussians, each defined by a center $\boldsymbol{\mu}_i$, covariance $\boldsymbol{\Sigma}_i$, opacity $\sigma_i$, and color $c_i$. Through differentiable splatting, 3DGS achieves real-time photorealistic rendering while maintaining an explicit representation~\cite{kerbl20233d}. With 3DGS also being a point cloud, common encoders for point clouds are utilized for 3D-VLM tasks. On the other hand, TRELLIS~\cite{xiang2025structured} can encode and decode Gaussian splats into an embedding space suitable for 3D content generation.

\subsection{Geometric View of Contrastive Alignment}
\label{subsec:contrastive-alignment}


Geometry, images, and language may describe the same object while remaining structurally incomparable. For instance, a~\textit{``chair''} can be represented by a point cloud, an RGB rendering, or a textual phrase~\textit{``a wooden chair''}. Their encoders initially produce features in modality-specific spaces, where distances are dominated by the properties of each encoder rather than by semantic correspondence.


Contrastive learning (CL) addresses this mismatch by mapping all modalities into a shared embedding space organized according to semantic compatibility. 
An intuitive analogy is to view the embedding space as a semantic map: \textit{representations describing the same object should occupy nearby locations, whereas representations of unrelated objects should be placed farther apart}.
%
This organization enables direct cross-modal comparison, including text-to-3D retrieval, image-to-3D retrieval, and zero-shot classification.

%

\begin{figure}[ht]
    \centering
    \vspace{-0.2cm}
    \includegraphics[width=0.99\linewidth]{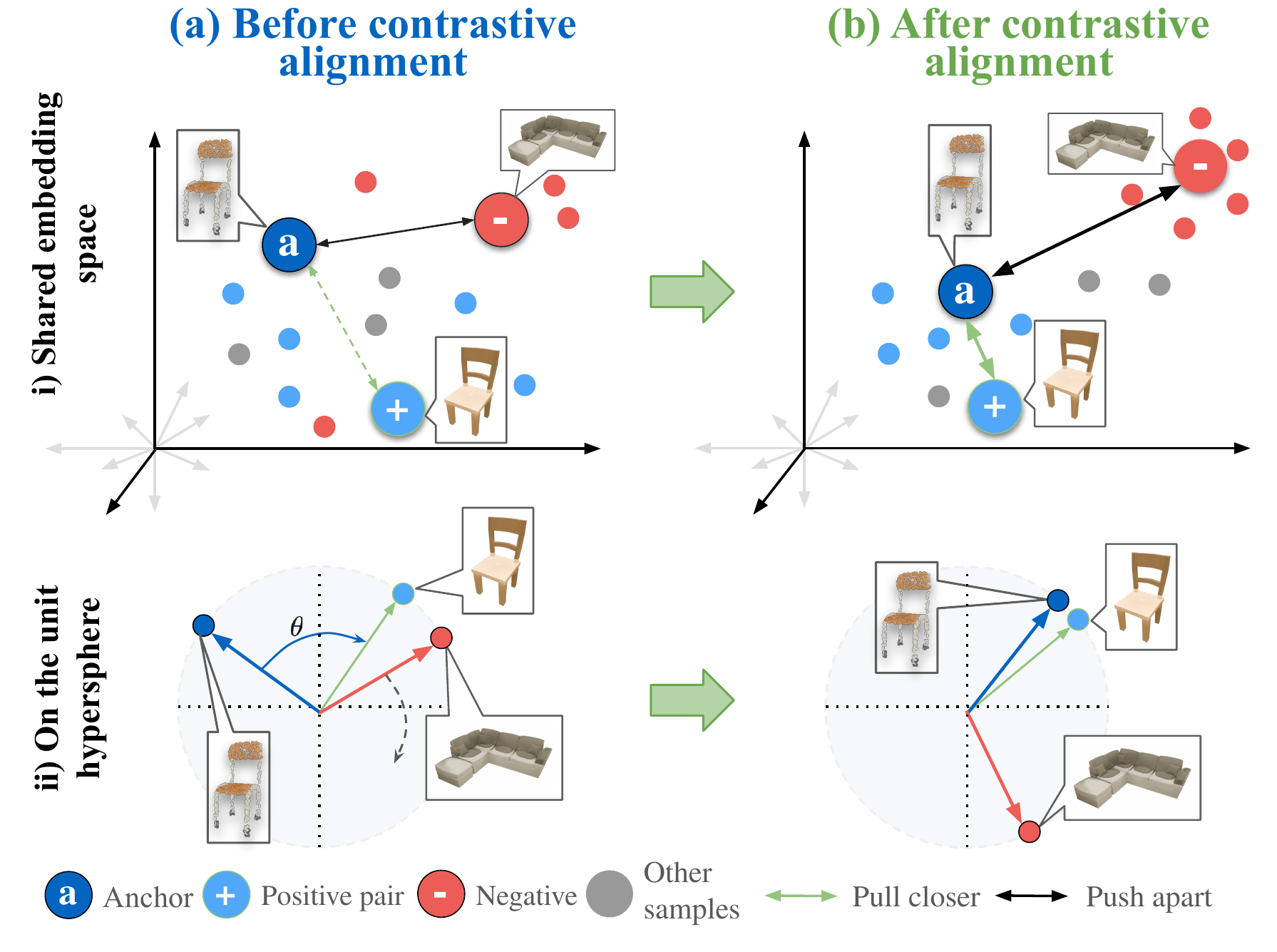}
    \caption{\textbf{Geometric intuition of contrastive alignment.} 
    The top panels visualize the shared embedding space, and the bottom panels show its angular interpretation on the unit hypersphere. 
    (a) Before alignment, a 3D anchor may be closer to a negative image sample than to its positive pair. 
    (b) Contrastive optimization changes this relation by increasing positive similarity relative to the remaining in-batch candidates.
    }
    \label{fig:cl-alignment}
        \vspace{-0.2cm}
\end{figure}

As shown in \autoref{fig:cl-alignment}, a 3D anchor (\circleanchor{a}) may initially be more similar to a negative candidate (\circlenegative{-}) than to its paired image/text (\circlepositive{+}). Contrastive optimization increases positive similarity relative to the remaining batch candidates. Since the embeddings are $\ell_2$-normalized, they lie on the unit sphere, where the inner product corresponds to cosine similarity; alignment thus reduces the angular distance $\theta$ between representations. Positive and negative denote correspondence relative to the anchor: the paired image or text is positive, whereas non-matching batch samples serve as negative candidates.

\subsection{Contrastive Alignment for 3D VLMs}
\label{subsec:alignment-vlms}
%

Here, we present the cross-modal alignment illustrated in \autoref{fig:cl-alignment}.
Let $M_a,M_b\in\{S,I,T\}$ denote 3D, image, or text modalities. 
Given $\ell_2$-normalized embeddings $z_i^{M_a}$ and $z_j^{M_b}$, their temperature-scaled cosine similarity is defined as  $s_{ij}^{a,b}=(z_i^{M_a})^{\top}z_j^{M_b}/\tau$, where $\tau>0$ controls the concentration of the similarity distribution.
\begin{figure}
    \centering
    \includegraphics[width=0.9\linewidth]{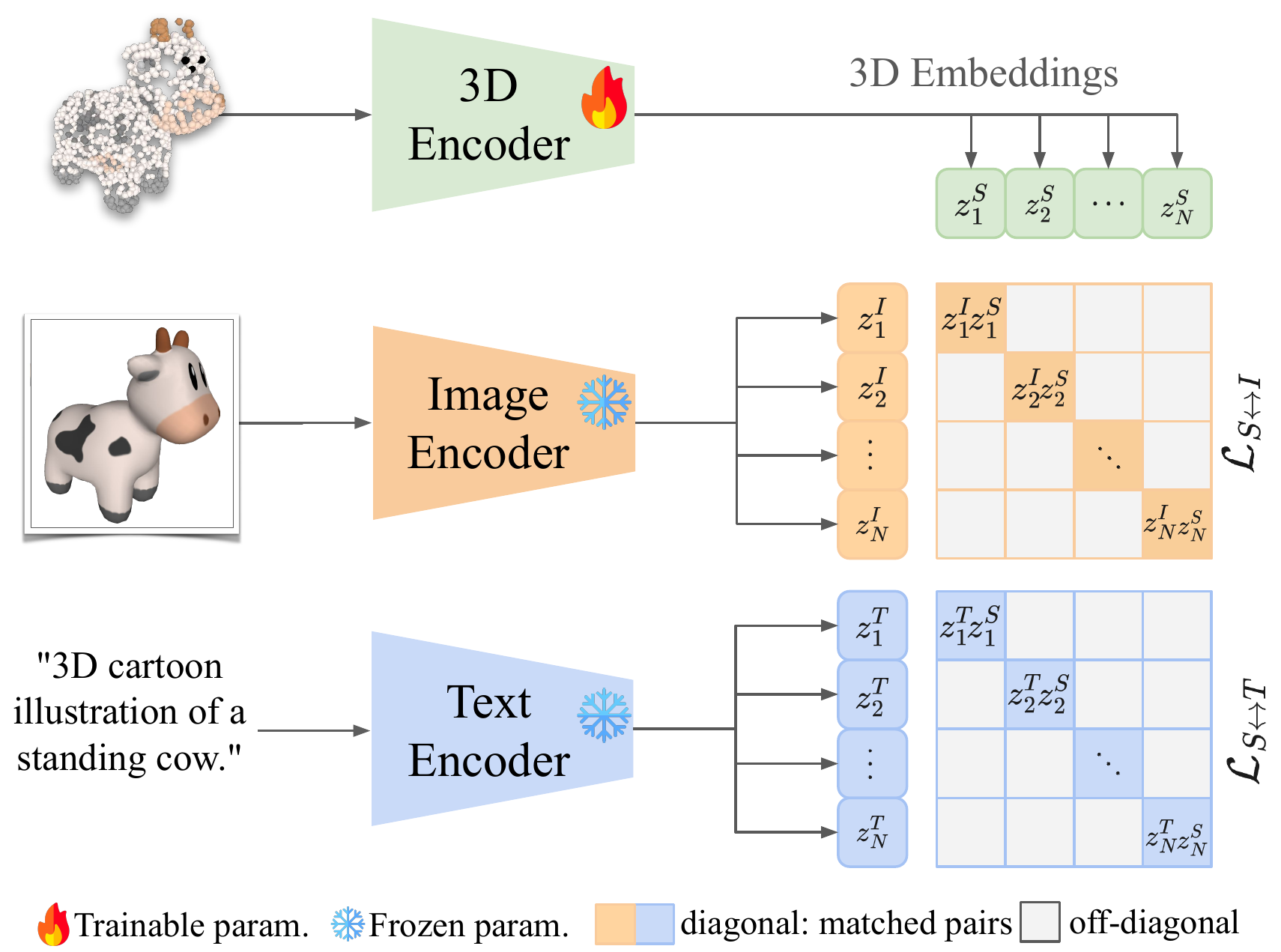}
    \vspace{4pt}
    \caption{\textbf{CLIP-style alignment for 3D VLMs.}~A 3D encoder aligns its embeddings with image and text encoders through 3D-Image and 3D-text similarity matrices. The main diagonal entries define matched positive pairs, whereas off-diagonal entries define the remaining negative candidates for symmetric InfoNCE~\cite{oord2018infonce} loss.}
    \label{fig:vlm-intuition}
    \vspace{-0.5cm}
\end{figure}
\autoref{fig:vlm-intuition} shows the corresponding CLIP-based pretraining pipeline. For a batch of $N$ matched triplets, the encoders produce 3D, image, and text embeddings $\{z_i^{S},z_i^{I},z_i^{T}\}_{i=1}^{N}$. Pairwise comparisons form the 3D--image and 3D--text similarity matrices. Their diagonal entries $(i=j)$ correspond to matched positive pairs, whereas off-diagonal entries $(i\neq j)$ serve as in-batch negative candidates.

For an anchor from modality $M_a$, the directional InfoNCE~\cite{oord2018infonce} objective identifies its paired representation among all candidates from modality $M_b$:
\begin{equation}
    \mathcal{L}_{M_a\rightarrow M_b}
    =
    -\frac{1}{\mathcal{N}}
    \sum_{i=1}^{\mathcal{N}}
    \log
    \frac{
        \exp\left(s_{ii}^{a,b}\right)
    }{
        \displaystyle
        \sum_{j=1}^{\mathcal{N}}
        \exp\left(s_{ij}^{a,b}\right)
    }.
    \label{eq:directional_infonce}
\end{equation}
Equation~\ref{eq:directional_infonce} can be interpreted as an $N$-way cross-modal matching objective: it increases the similarity of the paired representation relative to the remaining candidates. Following CLIP~\cite{Radford2021learning}, matching is performed in both directions:
\begin{equation}
    \mathcal{L}_{M_a\leftrightarrow M_b} =
    \frac{1}{2}
    \left( \mathcal{L}_{M_a\rightarrow M_b}
    +
    \mathcal{L}_{M_b\rightarrow M_a}
    \right).
\end{equation}

Tri-modal 3D VLMs commonly align 3D representation independently with both images and text:
\begin{equation}
\mathcal{L}_{\mathrm{3D\text{-}VLM}}
=
\mathcal{L}_{S\leftrightarrow I}
+
\mathcal{L}_{S\leftrightarrow T}.
\end{equation}

When the image and text encoders are inherited from CLIP~\cite{Radford2021learning}, this objective transfers their visual-semantic structure to the 3D encoder, enabling zero-shot recognition and cross-modal retrieval. Existing methods differ mainly in their 3D representation, backbone, supervision, pair construction, and trainable components~\cite{gao2024mixcon3d}.

\section{A Taxonomy of 3D Vision-Language Models}
\label{sec:taxonomy-vlms}

We organize this evolution into five main paradigms: (i) closed-set 3D recognition, (ii) projection-based VLM adaptation, (iii) native joint embedding models, (iv) language-aligned 3DGS representations, and (v) 3D VLLMs. These paradigms progressively replace fixed task-specific outputs with open-vocabulary representations and, recently, with generative language interfaces.

\subsection{From Closed-set 3D Recognition to Language Alignment}
\label{subsec:closed-set-3drecognition}
Early 3D learning followed a closed-set paradigm, in which geometric inputs were mapped to categories from a fixed taxonomy. MVCNN~\cite{su2015multi} represented objects through multiple rendered views, whereas PointNet~\cite{Qi2017pointnet} directly processed unordered point sets. Subsequent models improved geometric encoding through dynamic local graphs~\cite{wang2019dynamic} and transformer-based pretraining~\cite{pang2022masked}. Nevertheless, their output spaces remained defined by task-specific labels, preventing direct association with arbitrary natural-language concepts. Vision-language pretraining replaces the fixed classifier with a continuous semantic space.

\subsection{Projection-Based 3D VLM Adaptation}
\label{subesc:projection-adaptation-vlm}
%

%

Pretrained VLMs cannot directly process irregular 3D structures. Projection-based methods address this limitation by converting geometry into rendered views that can be processed by frozen 2D encoders. PointCLIP~\cite{Zhang2022PointCLIP} renders point clouds as multi-view depth maps and compares their CLIP features with
textual category embeddings. PointCLIP V2~\cite{Zhu2023PointCLIPV2} improves this transfer through more realistic projections and 3D-specific prompts, while EPCL~\cite{huang2024frozenclip} further explores frozen CLIP~\cite{Radford2021learning} features for point cloud understanding.

This paradigm provides a simple interface between 3D data
and pretrained 2D models, often without learning a dedicated
3D encoder. However, its performance depends on viewpoint
selection, rendering quality, and the geometric information
preserved by the projections.

\subsection{Joint Embedding Spaces}
\label{subsec:joint-embedding-vlms}
Joint-embedding methods instead train a dedicated 3D
encoder to produce features compatible with pretrained image
and text representations. As shown in Section~\ref{subsec:alignment-vlms}, given matched 3D, image, and natural language, contrastive learning organizes the three modalities within a shared semantic space while preserving the native geometric input.

CG3D~\cite{hegde2023clip} and CLIP2Point~\cite{Huang2023clip2point} explored early ways of 2D-to-3D transfer. ULIP~\cite{Xue2023ulip} established the widely adopted tri-modal formulation, aligning a point cloud encoder with frozen image and text encoders through 3D--image and 3D--text contrastive objectives.
Succeeding methods advanced this paradigm through data, supervision, and architectural scaling. OpenShape~\cite{Liu2023openshape} emphasized the use of large and diverse 3D datasets for open-world recognition, such as Objaverse~\cite{Deitke2023objaverse, Deitke2023objaversexl}. Uni3D~\cite{Zhou2023uni3d} adapted scalable ViT backbones to point clouds, while ULIP-2~\cite{Xue2024ulip} replaced manual captions with automatically generated descriptions. Other approaches improve multimodal augmentation~\cite{gao2024mixcon3d}, supervision~\cite{mao2024opendlign, lobo20263dmrl}, modality-gap modeling~\cite{lee2025duoduo} or lightweight adapters~\cite{Zhang2024tamm}.

More recent work moves toward general-purpose spatial encoders beyond joint embedding alignment.
Concerto~\cite{zhang2025concerto} combines 3D self-distillation with 2D--3D joint embedding and maps its representations into CLIP's language space through a linear translator.
Utonia~\cite{zhang2026utonia} extends this direction by learning a unified self-supervised encoder across heterogeneous point cloud domains, whose features also benefit multimodal spatial reasoning and embodied policies.

However, these methods preserve more geometric information than projection-based methods and have become a central paradigm for embedding-based 3D VLMs. The limitations include the need for large paired datasets, limited or noisy multimodal supervision, and the cost of contrastive pre-training.



\subsection{3D Gaussian Splatting Meets VLMs}
\label{subsec:3dgs_vlms}

Point clouds explicitly represent geometry but offer limited appearance information. As mentioned in Section~\ref{subsec:3d-representations}, 3DGS represents a scene through renderable primitives that jointly encode position, covariance, opacity, and color~\cite{kerbl20233d, matias2025volume, perazzo2025study}. This provides a direct connection between 3D structure and the 2D images processed by VLMs.

UniGS~\cite{li2025unigs} extends tri-modal alignment to Gaussian representations by matching Gaussian-aware features with image and language embeddings.
TIGaussian~\cite{toptigaussian} decomposes Gaussian attributes into specialized latent branches and introduces multi-view fusion for spatially coherent alignment.
CLIP-GS~\cite{CLIPGS2025} focuses on efficient Gaussian tokenization and viewpoint-aware supervision for the cross-modal retrieval task.

When compared with point clouds, Gaussian representations provide a closer coupling between geometry, appearance, and image supervision. However, they remain dependent on reconstruction quality and can introduce substantial storage or per-scene costs.

\subsection{3D Vision Large Language Models}
\label{subsec:3dvllms}

Prior 3D VLMs primarily support similarity-based tasks such as recognition and retrieval. 3D VLLMs instead connect geometric representations to autoregressive language models, enabling captioning, question answering, instruction following, and spatial reasoning. Here, the architecture combines a 3D encoder or tokenizer, a learned projector, and a pretrained language backbone, such as Qwen3-VL~\cite{bai2025qwen3}.

PointLLM~\cite{xu2024pointllm} and ShapeLLM~\cite{qi2024shapellm} established a common architecture in which a 3D encoder extracts geometric features, a learned projector maps them into the LLM embedding space, and the language model generates responses conditioned on both 3D and textual instructions.
PointAlign~\cite{zhang2025pointalign} introduces feature regularization to preserve geometric information during language modeling.
Multi-3DLLM~\cite{li2025beyond} extends reasoning to multi-object relations, while SAGE~\cite{huang2025sage} directly converts point clouds into discrete tokens. N3D-VLM~\cite{cheng2025n3dvlm} further combines native 3D grounding with language-based spatial reasoning.
This transition from 3D VLMS to 3D VLLMs changes the learning objective from cross-modal comparison to language generation. This enables interaction with 3D content but introduces challenges related to hallucination, geometric fidelity, token efficiency, and limited 3D instruction data.

\section{Current Trends and Emerging Applications}
\label{sec:tasks}
\begin{figure}[htpb]
    \centering
     \vspace{-0.2cm}
    \includegraphics[width=1\linewidth]{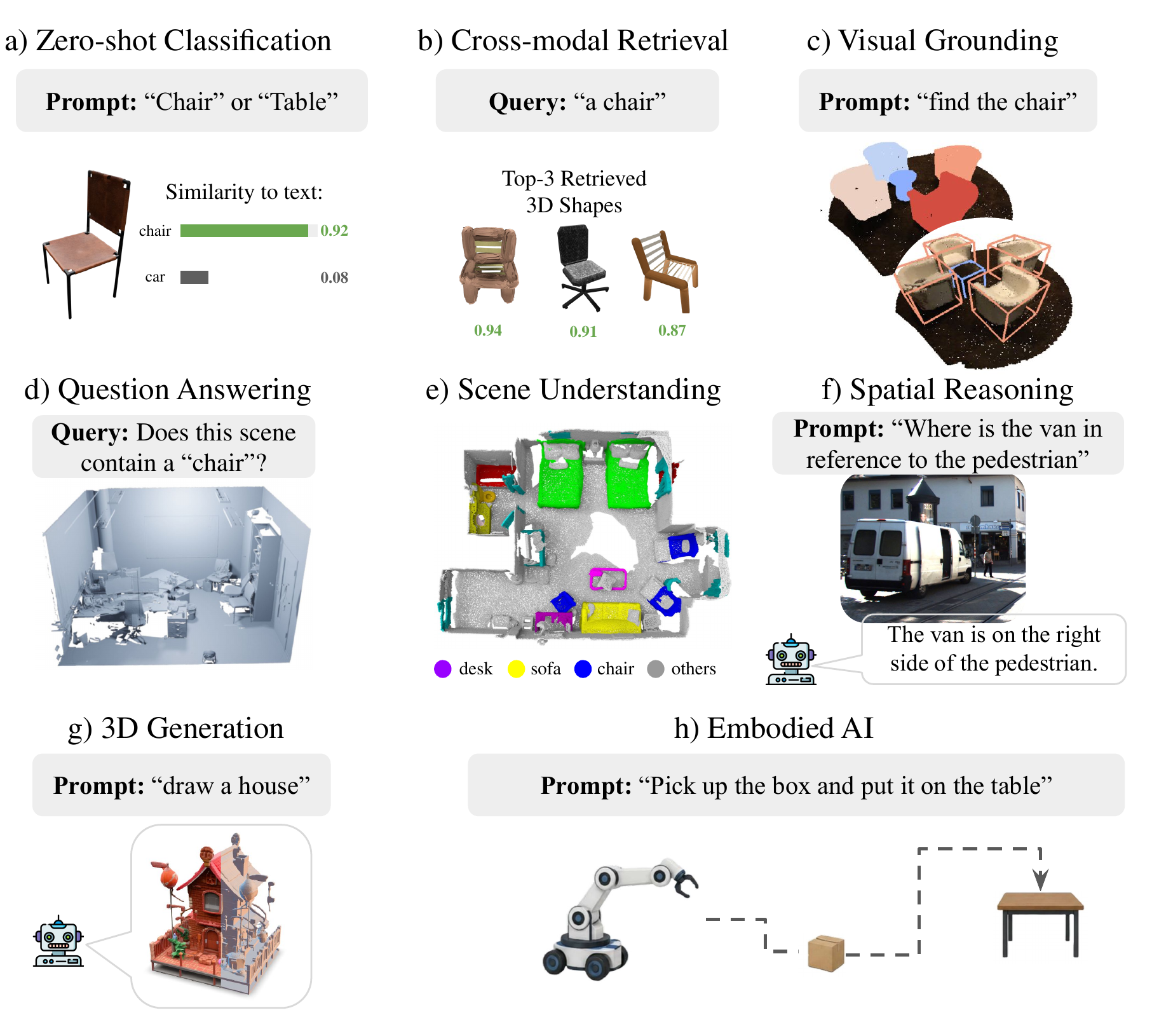}
    \caption{\textbf{Main tasks and applications of 3D VLMs.}}
    \label{fig:tasks}
    \vspace{-0.3cm}
\end{figure}

Recent advances in 3D representations and multimodal foundation models allow 3D VLMs to also support scene understanding, spatial reasoning, content generation, and embodied interaction. We discuss five emerging directions shaping the field, as shown in~\autoref{fig:tasks}.
\subsection{Recognition, grounding and retrieval}
At the object level, \emph{zero-shot classification} matches 3D representations with textual category descriptions (Figure~\ref{fig:tasks}(a)). PointCLIP~\cite{Zhang2022PointCLIP} and PointCLIP V2~\cite{Zhu2023PointCLIPV2} transfer CLIP through multi-view projections, whereas ULIP~\cite{Xue2023ulip}, OpenShape~\cite{Liu2023openshape}, and Uni3D~\cite{Zhou2023uni3d} learn native language-aligned 3D embeddings. The same shared embedding space supports \emph{cross-modal retrieval} (Figure~\ref{fig:tasks}(b)). ULIP and OpenShape retrieve 3D shapes from text or image queries, while CLIP-GS~\cite{CLIPGS2025} extends retrieval to Gaussian representations. 
\emph{3D visual grounding} extends recognition to spatial localization (Figure~\ref{fig:tasks}(c)). Multi-3DLLM~\cite{li2025beyond} models multi-object relations, while N3D-VLM~\cite{cheng2025n3dvlm} preserves native geometric features for language-guided grounding and spatial reasoning. Related open-vocabulary methods localize objects and parts specified only at inference time. These tasks therefore differ mainly in their outputs: category scores, spatial regions, or ranked 3D assets.

\subsection{Scene Understanding and Spatial Reasoning}

Beyond individual objects, 3D VLLMs enable language-based interaction with complete scenes. For \emph{3D question answering} (Figure~\ref{fig:tasks}(d)), PointLLM~\cite{xu2024pointllm} and ShapeLLM~\cite{qi2024shapellm} reason about object properties, while Multi-3DLLM~\cite{li2025beyond} extends this capability to multi-object relations.
For \emph{scene understanding} (Figure~\ref{fig:tasks}(e)), language-aligned Gaussian methods embed semantics directly into explicit 3D scenes. LEGaussians~\cite{shi2024language} reduces multi-view inconsistencies, whereas Gaussian Grouping~\cite{ye2024gaussian} and Unified-Lift~\cite{zhu2025rethinking} support instance segmentation, grouping, and editing. OpenGaussian~\cite{wu2024opengaussian} enables point-level open-vocabulary understanding, GaussianCut~\cite{jain2024gaussiancut} provides prompt-driven object isolation, and SceneSplat~\cite{li2025scenesplat} predicts language-aligned Gaussian features without per-scene optimization.
These representations further enable \emph{spatial reasoning} (Figure~\ref{fig:tasks}(f)). Multi-3DLLM~\cite{li2025beyond} models relations among multiple objects, while N3D-VLM~\cite{cheng2025n3dvlm} preserves native 3D grounding for more accurate geometric reasoning. Interactive methods such as GaussianCut~\cite{jain2024gaussiancut} and SAM~3D~\cite{chen2025sam3d} additionally allow scenes to be queried or manipulated through text, clicks, or spatial prompts.

\subsection{3D Generation}
\label{subsec:generation}

Multimodal models also condition the synthesis of geometry and appearance (Figure~\ref{fig:tasks}(g)). MeshGPT~\cite{siddiqui2024meshgpt} and recent MeshFlow variants~\cite{li2026meshflow,sun2026meshflow} investigate autoregressive or flow-based geometric mesh generation, but they are geometric generators rather than VLMs. 
TRELLIS~\cite{xiang2025structured}, in contrast, conditions structured 3D generation on text or image prompts and supports multiple output formats, including meshes, radiance fields, and 3D Gaussians. SAM~3D~\cite{chen2025sam3d} addresses visually grounded single-image reconstruction, predicting geometry, texture, and layout from an input image. These methods illustrate complementary advances in geometric generation, multimodal conditioning, and image-based 3D reconstruction.

\subsection{Embodied AI}
\label{subsec:embodied-ai}
The final direction connects multimodal 3D understanding to physical interaction (Figure~\ref{fig:tasks}(h)). CLIPort~\cite{shridhar2022cliport} combines semantic and spatial pathways for robotic manipulation, while Gato~\cite{reed2022generalistagent} introduced a generalist sequence model across perception and control tasks. More recent Vision-Language-Action (VLA) models, including RT-2~\cite{zitkovich2023rt}, Octo~\cite{team2024octo}, OpenVLA~\cite{kim2024openvla}, and GR00T N1~\cite{bjorck2025gr00t}, map visual observations and language instructions directly to robot actions.
Most of these systems rely primarily on 2D observations. Explicit 3D representations can complement them with metric depth, object extent, and spatial relations. Remaining challenges include action latency, limited 3D--language--action data, robustness to unseen
environments, and preservation of geometric accuracy during control~\cite{landauvlaambiguity}.




\section{Conclusion and Future Directions}

This tutorial has explored the integration of 3D representations within VLMs, tracing the evolution from 2D projections to native joint embeddings, 3DGS, and 3D VLLMs.
Despite recent progress, unifying 3D data and natural language remains challenging due to the lack of scalable encoders that generalize across point clouds, meshes, Gaussian primitives, and structured 3D latents. Current Embodied VLA models also make limited use of explicit geometry, restricting metric and spatial reasoning during physical interaction. Future work should focus on efficient multimodal alignment, scalable 3D-language-action datasets, and robust transfer from synthetic assets to real-world sensor data. These advances may establish 3D VLMs as a unified interface between language, 3D environments, and embodied agents.





\bibliographystyle{backend/IEEEtran}
\bibliography{backend/references}

\end{document}